\documentclass{article}

\PassOptionsToPackage{numbers, compress}{natbib}
\usepackage[final]{neurips_2023}
\usepackage{natbib}
\usepackage{algorithm}
\usepackage{xcolor}
\usepackage{amssymb}
\usepackage{algpseudocode}
\usepackage{tcolorbox}
\usepackage{subfigure}
\usepackage{graphicx}
\usepackage{textcomp}
\usepackage{xcolor}
\usepackage{wrapfig}
\usepackage{bm}

\usepackage[utf8]{inputenc} 
\usepackage[T1]{fontenc}    
\usepackage{hyperref}       
\usepackage{url}            
\usepackage{booktabs}       
\usepackage{amsmath}
\usepackage{nicefrac}       
\usepackage{microtype}      
\usepackage{xcolor}         

\title{Successfully Applying Lottery Ticket Hypothesis to Diffusion Model}

\author{%
  Chao Jiang\textsuperscript{\rm 1} Bo Hui\textsuperscript{\rm 2}\thanks{Corresponding author}\:\: Bohan Liu\textsuperscript{\rm 3} Da Yan\textsuperscript{\rm 4}\\
  Visa\textsuperscript{\rm 1}, University of Tulsa\textsuperscript{\rm 2}, Carnegie Mellon University\textsuperscript{\rm 3}, Indiana University Bloomington\textsuperscript{\rm 4}\\
  \texttt{chajiang@visa.com, bo-hui@utulsa.edu} \\
  \texttt{bohanli2@andrew.cmu.edu, yanda@iu.edu}
}

\begin{document}
\maketitle
\begin{abstract}
 Despite the success of diffusion models, the
training and inference of diffusion models are notoriously expensive due to the long chain of the reverse process. In parallel, the Lottery Ticket Hypothesis (LTH) claims that there exists winning tickets (i.e., a
properly pruned sub-network together with original weight initialization) that can
achieve performance competitive to the original dense neural network when trained in isolation. In this work, we for the first time apply LTH to diffusion models. We empirically find subnetworks at sparsity $90\%-99\%$ without compromising
performance for denoising diffusion probabilistic models on benchmarks (CIFAR-10, CIFAR-100, MNIST). Moreover, existing LTH works identify the subnetworks with a unified sparsity along different layers. We observe that the similarity between two winning tickets of a model varies from block to block. Specifically, the upstream layers from two winning tickets for a model tend to be more similar than the downstream layers. Therefore, we propose to find the winning ticket with varying sparsity along different layers in the model. Experimental
results demonstrate that our method can find sparser sub-models that require less memory for storage
and reduce the necessary number of FLOPs. Codes are available at~\url{https://github.com/osier0524/Lottery-Ticket-to-DDPM}.
\end{abstract}

\section{Introduction}

Diffusion models~\cite{DBLP:conf/icml/Sohl-DicksteinW15, DBLP:conf/nips/HoJA20, DBLP:conf/iclr/0011SKKEP21} have achieved state-of-the-art results in a wide range of applications such as image generation~\cite{DBLP:conf/iclr/SongME21, DBLP:conf/icml/NicholD21, DBLP:conf/nips/KarrasAAL22, DBLP:conf/cvpr/RombachBLEO22}, text-to-image~\cite{DBLP:conf/nips/SahariaCSLWDGLA22, DBLP:conf/icml/NicholDRSMMSC22, DBLP:journals/corr/abs-2204-06125}, video generation~\cite{DBLP:journals/corr/abs-2203-09481, DBLP:conf/nips/HoSGC0F22}, audio generation~\cite{DBLP:conf/iclr/KongPHZC21, DBLP:conf/icml/PopovVGSK21}, and protein generation~\cite{DBLP:conf/icml/YimTBMDBJ23, DBLP:conf/iclr/TrippeYTBBBJ23}. These generative models are powerful to produce high-quality data by corrupting the
data with slowly increasing noise and then learning to reverse this corruption. For example, Denoising diffusion probabilistic modeling (DDPM)~\cite{DBLP:conf/nips/HoJA20} trains a sequence of probabilistic models to reverse each step of the noise corruption.

Although diffusion models have shown impressive performance in capturing distributions and sample quality, they
are notoriously slow to generate data due to the long chain of reversing the diffusion process. The noisy data will go through the same U-Net-based generator network thousands
of times or even more~\cite{DBLP:conf/nips/HoJA20, DBLP:conf/iclr/WangZHCZ23}. At the same time, diffusion models are also notoriously hungry to train.
They
require many iterations and large size of data to capture the complex data distributions. For example, it takes over two weeks
to train DDPM~\cite{DBLP:conf/nips/HoJA20} on eight V100 GPUs for $256 \times 256$ resolution datasets. The reported training time of a
state-of-the-art diffusion model in~\cite{DBLP:conf/nips/DhariwalN21} is over 100 days on V100 GPU
days to generate high-quality image samples. Moreover, as the image resolution
and the size of the training data increases, the training and inference costs grow exponentially. To improve the training efficiency and inference speed, many efficient sampling methods have been proposed, such as DDIM~\cite{DBLP:conf/iclr/SongME21}, DPM-Solver~\cite{DBLP:conf/nips/0011ZB0L022}, EDM-Sampling~\cite{DBLP:conf/nips/KarrasAAL22}. Different from these fast solvers, we propose to mitigate the computational cost by pruning the reverse model.

The Lottery Ticket Hypothesis (LTH)~\cite{DBLP:conf/iclr/FrankleC19} states that a dense neural network model contains a highly sparse subnetwork (i.e., winning tickets) that can achieve even better performance than the original model. The winning tickets can be identified by training a network and pruning its
parameters with the smallest magnitude in an iterative way or one-shot way. Before it is trained in each iteration, the weights will be reset to the original initialization. The identified winning tickets are retrainable to reduce the high memory cost and long inference time of the original neural networks, which has been proved by many works~\cite{DBLP:journals/corr/abs-1903-01611, DBLP:conf/iclr/DiffenderferK21, DBLP:conf/icml/MalachYSS20, DBLP:conf/nips/ZhouLLY19, DBLP:conf/icml/ChenCMWW22}. The existence of winning tickets has been verified in both experiments~\cite {DBLP:conf/nips/MaYSCCCLQLWW21,DBLP:conf/nips/SuCCWG0L20} and theory~\cite{DBLP:conf/nips/ZhangJZZZRLWJD21,DBLP:conf/nips/SakamotoS22, DBLP:conf/iclr/BurkholzLMG22, DBLP:conf/iclr/FerbachTGB23,DBLP:conf/icml/MalachYSS20, DBLP:conf/iclr/CunhaNV22}.  LTH has
been extended to find the winning tickets for different kinds of neural networks such as GANs~\cite{DBLP:conf/iclr/ChenZSC21, DBLP:conf/aaai/KalibhatBF21}, Transformers~\cite{DBLP:conf/acl/BrixBN20, DBLP:conf/emnlp/PrasannaRR20, DBLP:conf/nips/ChenFC0ZWC20, DBLP:conf/emnlp/BehnkeH20} and GNNs~\cite{DBLP:conf/icml/ChenSCZW21, DBLP:conf/iclr/Hui0MK23, DBLP:conf/bigdataconf/HarnYHZSSK22}.
It has also been applied in various domains including computer vision and natural language processing~\cite{DBLP:conf/aaai/GanCLC0WLW022, DBLP:conf/acl/ZhengRZLWWGZH22, DBLP:conf/cvpr/ChenFC0ZCW21}. We for the first time propose to apply LTH to diffusion models. Different from existing works based on efficient sampling, we aim to reduce the number of parameters for efficient training and inference. Specifically, we perform the empirical study to investigate
whether there exists a trainable subnetwork of the diffusion model with original initialization that can achieve competitive performance than the original diffusion model. The answer is affirmative. We conclude that the winning tickets achieve the same performance with $90\%$ floating-point operations (FLOPs) saving on the original diffusion model. 

We remark that the existing works in LTH identify the winning ticket with a unified sparsity along different layers. That is, we use the same pruning ratio to mask the parameters in the model. In this paper, we empirically found that the similarity between two winning tickets of a given model varies from module to module. Specifically, we introduce centered kernel alignment (CKA) as an index to measure the similarity between the sparsified modules from two winning tickets. We observe that the similarity at the upstream modules is higher than that at the downstream modules. This motivates us to configure the pruning ratio to be different at different modules. Based on the observation, we configure the pruning ratio to be lower at the upstream layer so that the sparsity will be lower for these layers. Intuitively, we need to make sure there are enough parameters to be trained so that meaningful full hidden states can be learned from noisy input data. In the experiment, we verify that our configuration can result in sparser winning tickets without performance compromise.

Since the combination of sparse architectures and initializations in a 
 winning ticket can reveal the potential implications for theoretical
study of optimization and generalization in diffusion models, we can take inspiration from winning tickets to design new
architectures for the diffusion process, we hope to stimulate the research
progress of improving the inference speed of diffusion models.

The contribution of this work can be summarised as:
\vspace{-2mm}
\begin{itemize}
    \item We for the first time apply the lottery ticket hypothesis to the diffusion model. Using a pruning method based on magnitude, we identify subnetworks at 99\% sparsity in
DDMP without performance compromise.  
    \item We propose to identify a winning ticket with a varied sparsity along different layers, which is different from existing pruning algorithms in LTH. The proposed method can result in winning tickets with higher sparsity.
    \item The empirical result verifies the quality of pictures generated by a winning ticket is even higher than that generated by the original DDPM.
\vspace{-3mm}
\end{itemize}
\section{Preliminary}
\vspace{-2mm}
We focus on the DDPM in this paper. Given an input $\bf x_0$, the diffusion process gradually adds Gaussian noise based on a variance schedule $\beta_1, \cdot, \beta_T$. Denote $\bm\theta$ as the parameters to learn the distribution $p_{\bm\theta}(\bf x_{t-1}|\bf x_t)=\mathcal{N}(\bf x_{t-1}; \bm{\mu}_{\bm{\theta}}(\bf x_t, t),\sum_{\bm\theta}(\bf x_t, t))$ in the reverse process.

Given the neural network parameterized by $\bm\theta$, a subnetwork is parameterized by $\bm\theta\odot\bm m$, where $\bf m\in \{0,1\}^{||\bm\theta||_0}$
is a pruning mask for $\bm\theta$ and $\odot$ indicates the element-wise product. We use $||\cdot||_0$ to represent the $L_0$ norm counting the number of non-zero elements. The value $0$ in $\bf m$ means the corresponding parameter $\bm\theta$ will be masked. The sparsity of a subnetwork is measured as $1-\frac{||\bf m ||_0}{||\bm\theta||_0}$.

Modern pruning methods can be classified into structured pruning and unstructured pruning. In general, structured pruning removes entire groups of neurons, filters, or channels of neural networks while unstructured pruning results in unstructured sparse matrices. The pruning method based on the magnitude in LTH belongs to the unstructured pruning category. We use an iterative way to find a subnetwork $\bm\theta_\tau\odot\bm m$, where $\bm\theta_\tau$ is
the rewound initialization, which can reach the comparable performance to the full network within a similar
training iteration when trained in isolation.  After each iteration of training and pruning, we rewind the model with the parameters at $\tau$ epoch. The combination of $\bm\theta_\tau$ and $\bf m$ with comparable performance is defined as a winning ticket.
\vspace{-3mm}
\section{The Existence of Winning Tickets in DDPM}
\vspace{-2mm}
Existing works find the winning ticket by pruning the smallest magnitude in an iterative way.
Given a pruning ratio $p\%$, we will sort the magnitude of weights after training and prune $p\%$ of parameters with the lowest magnitudes. In practice, existing work prunes the model layer by layer. It will result in a subnetwork where all layers have the same sparsity ($p\%$).

We remark that it is not necessary to find a subnetwork with a unified sparsity along different layers. Intuitively, the input data is highly noisy in the denoising process and we need more parameters to learn a meaningful full hidden state. In this paper, we measure the similarity of two winning tickets based on canonical correlation analysis. Let $ W^1_{i}$ and  $W^2_{i}$ be the sparsified weight matrix of $i$th layer in the first and second winning ticket. We introduce Hilbert-Schmidt Independence Criterion (HSIC) to 
measure the similarity ~\cite{DBLP:conf/alt/GrettonBSS05} :
$\textrm{HSIC}(K, L)=\frac{1}{(n-1)^2}{tr}(KHLH)$
where $K_{i,j,k}=k(W^1_{i,j}, W^1_{i,k})$ and $L_{i, j,k}=l(W^2_{i,j}, W^2_{i,k})$, and $H$ is the centering matrix. Both $k(\cdot)$ and $l(\cdot)$ are the RBF kernels.  HSIC
can be considered as the maximum mean discrepancy between the
joint distribution and the product of the marginal distributions~\cite{DBLP:conf/icml/Kornblith0LH19}. The normalized similarity index is defined as:
$$
    \textrm{CKA}(K, L) = \frac{\textrm{HSIC}(K, L)}{
\textrm{HSIC}(K, K)\textrm{HSIC}(L, L)}.
$$

\begin{wrapfigure}{R}{0.3\textwidth}
\centering
\vspace{-4mm}
\includegraphics[width=\linewidth]{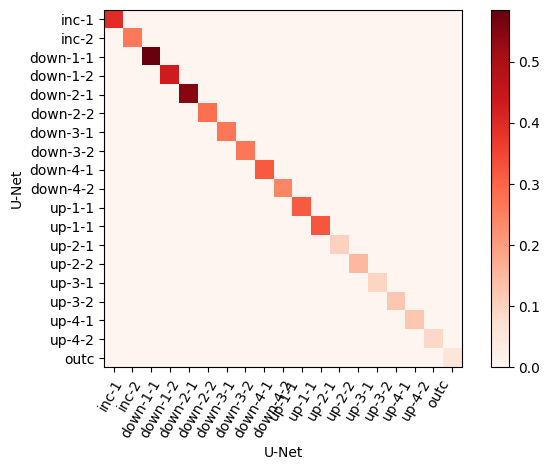}
\vspace{-7mm}
\caption{Similarity between two winning tickets} \label{fig:sim} 
\vspace{-2mm}
\end{wrapfigure}
The effectiveness of this index has been verified by~\cite{DBLP:conf/iclr/DavariHNLWB23}. In Figure~\ref{fig:sim}, we show the CKA between two "Conv2d" modules with the same order in the sequence from two different winning tickets of U-Net. We observe that the similarity is higher at upstream modules. It means there could be a potential implication at these upstream layers.

Motivated by this observation, we propose to configure the pruning ratio to be lower at the upstream modules so that the sparsity will be lower for these modules. Algorithm~\ref{alg: iterative} describes our pruning method. Take the U-Net model as an example. We first train the model parameter as $\bm\theta_i$ after $i$ iterations. Then for each of $J$ modules (conventional blocks in U-Net) in the sequential list, we prune the parameters with the lowest magnitude. We gradually increase the pruning ratio by $q$ as the index of a module increases in the U-Net model. After pruning, we will rewind the parameter to an early stage ($\tau=5\%\ast i$) for the next iteration of pruning. We repeat this process until the desired sparsity $\delta$ is reached.
\begin{algorithm}[h]

\caption{Finding winning tickets for DDPM}
\label{alg: iterative}
\textbf{Input}:{ Initial parameter $\bm\theta_0$, initial mask $\mathbf{m}=\mathbf{1}\in \mathbb {R}^{||\mathbf{\bm\theta||}}$, pruning ratio $p$, incremental rate $q$}\\
\textbf{Output}: {Sparsified masks $\mathbf{m}$}
\begin{algorithmic}[1]
\While {$1-\frac{\|\mathbf m|_0}{\|\bm\theta\|_0} < \delta$}
    \State {Train the diffusion model based on gradient $\nabla_{\bm \theta}$ for $i$ iterations}
    \State {Arrive at parameters $\bm \theta_i$}
    \For{ module $j = 0, 1, 2, \cdots, J-1$}
        \State {Pruning $(p+j\ast q)\%$ of the lowest-scored values in $j$th module $\bm\theta^{(j)}$}
        \State {creating mask $\bf m^{(j)}$} for $j$th module
      \EndFor
      \State {Rewinding parameters to $\bm\theta_\tau$}
    \State {$\bm m=\{\bf m^{(0)}, \bf m^{(1)},\cdot,\bf m^{(J-1)}\}$}
	
\EndWhile
\end{algorithmic}
\vspace{-1mm}
\end{algorithm}

\noindent\textbf{Open discussion.} We raise a new question regarding improving the efficiency of reversing: can we use sub-networks with different sparsity in the reverse process? Since a winning ticket can be considered an equivalent version of the original model, we can leverage a sparser sub-network in the late stage of denoising to further improve efficiency. Intuitively, the noise in the later reverse process has been reduced and it will be easier to optimize. The challenge lies in how to optimize a dense model and a winning ticket while guaranteeing the convergence of training. Another challenge is to decide at which step to use the winning ticket. We leave this open question for future investigation which we hope to stimulate the research on improving the efficiency of the diffusion model.
\vspace{-2mm}
\section{Experiment}
\begin{figure}[h]
\vspace{-3mm}
\centering
	\subfigure[CIFAR-10]
	{
\includegraphics[,width=0.315\linewidth]{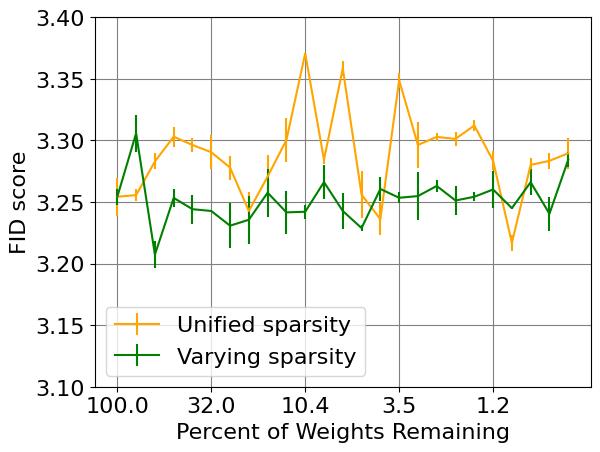}\vspace{-1cm}
}	
\subfigure[CIFAR-100]
	{
\includegraphics[trim=62 0 0 0,clip,width=0.27\linewidth]{cifar-10.png}\vspace{-1cm}
}
\subfigure[MNIST]
	{
\includegraphics[trim=62 0 0 0,clip,width=0.27\linewidth]{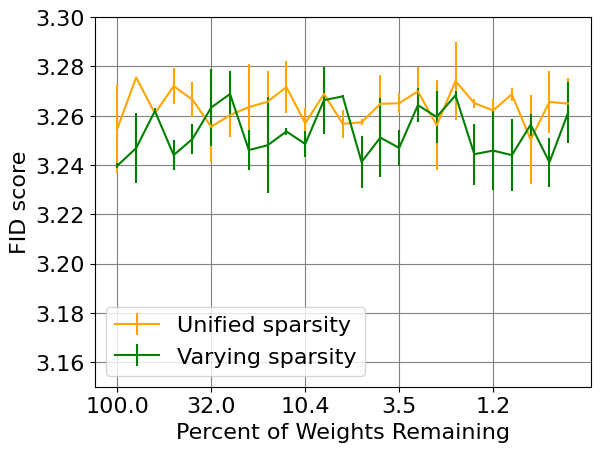}
\vspace{-3cm}
}
\label{fig: res}
\caption{ Performance of DDPM w.r.t. sparsity of Unet}
\vspace{-2mm}
\end{figure}

\vspace{-1mm}
We conducted experiments to find the winning tickets in DDPM. We use CIFAR-10, CIFAR-100, and MNIST as our
benchmark datasets. See the Appendix for more details of the experiment setting.
Figure~\ref{fig: res} shows the FID score with respect to the sparsity of the pruned U-Net. We observe that a sparsified model can even outperform the original model in terms of FID score. Moreover, by varying the sparsity, we can further reduce the sparsity of a winning ticket. The results verify the existence of winning tickets in DDPM, showing that we can find a winning ticket at sparsity $90\%-99\%$ on the three benchmark datasets.

\begin{figure}[h]
\centering
\vspace{-1mm}
\subfigure[Original DDPM]
	{
\includegraphics[width=0.8\linewidth]{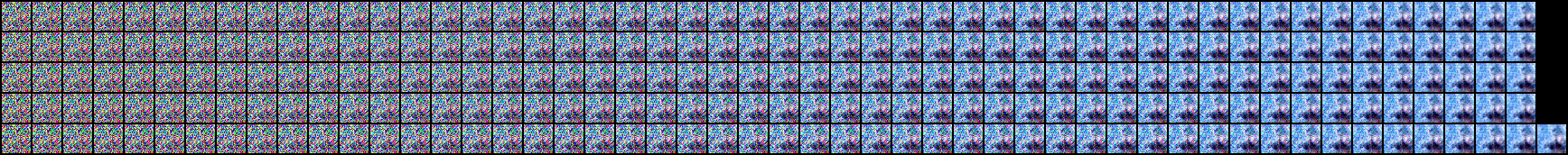}
}
\subfigure[Winning ticket (Sparsity: 99.4\%)]
	{
\includegraphics[width=0.8\linewidth]{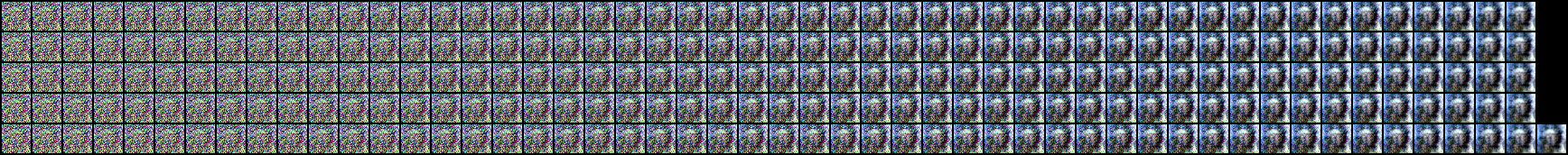}
}
\vspace{-2mm}
\caption{ Performance of DDPM w.r.t. sparsity of Unet}
\vspace{-2mm}
\label{fig: reverse}
\end{figure}

\begin{figure}[h]
\centering
\vspace{-3mm}

\subfigure[Original DDPM]
	{
\includegraphics[width=0.23\linewidth]{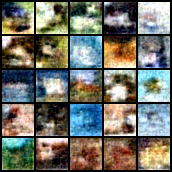}\vspace{-1cm}
}
\subfigure[Sub-network (Sparsity: 67.2\%]
	{
 
\includegraphics[width=0.23\linewidth]{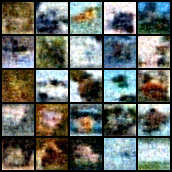}
\vspace{-1cm}
}
\subfigure[Winning ticket (Sparsity: 99.4\%))]
	{
 
\includegraphics[width=0.23\linewidth]{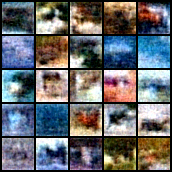}
\vspace{-1cm}
}
\vspace{-2mm}
\caption{\label{fig:acc2} Samples generated on CIFAR-10}
\vspace{-2mm}
\label{fig: set}
\end{figure}

We also visualize the quality of pictures generated by the winning ticket. Figure~\ref{fig: reverse} depicts the denoising process of both the winning ticket and the original model on the CIFAR-10 dataset. We can see that the quality of the generated picture is still high when the sparsity is 99.4\%. Figure~\ref{fig: set} shows samples generated by the winning tickets. It further verifies that a winning ticket can generate a picture with the same quality as the original model. We remark that we reduce 90\% of FLOPs with the winning ticket compared with the original model.


\bibliographystyle{plain}
\bibliography{neurips_2023}

\appendix
\section{Appendix}
We have introduced three benchmark datasets in the experiment: CIFAR-10, CIFAR-100, and MNIST. A U-Net model is used in the DDPM model. We ran the experiment on a machine with 8 NVIDIA Tesla A100 GPUs. All the training parameters (e.g., training epochs, time steps, and learning rate) are configured as the default of the original DDPM model. We Prune the model for 25 iterations. The default pruning ratio is 20\% and the incremental ratio is 1\% by default.
\end{document}